# Features based Mammogram Image Classification using Weighted Feature Support Vector Machine

Kavitha S
Department of Computer Science and Engineering
SSN College of Engineering, Chennai – 603110, India.
E-mail: kavithas@ssn.edu.in

Thyagharajan K K
Department of Information Technology
RMK College of Engineering and Technology, Chennai – 601206, India
E-mail: kkthyagharajan@yahoo.com

**Abstract**. In the existing research of mammogram image classification, either clinical data or image features of a specific type is considered along with the supervised classifiers such as Neural Network (NN) and Support Vector Machine (SVM). This paper considers automated classification of breast tissue type as benign or malignant using Weighted Feature Support Vector Machine (WFSVM) through constructing the precomputed kernel function by assigning more weight to relevant features using the principle of maximizing deviations. Initially, MIAS dataset of mammogram images is divided into training and test set, then the preprocessing techniques such as noise removal and background removal are applied to the input images and the Region of Interest (ROI) is identified. The statistical features and texture features are extracted from the ROI and the clinical features are obtained directly from the dataset. The extracted features of the training dataset are used to construct the weighted features and precomputed linear kernel for training the WFSVM, from which the training model file is created. Using this model file the kernel matrix of test samples is classified as benign or malignant. This analysis shows that the texture features have resulted in better accuracy than the other features with WFSVM and SVM. However, the number of support vectors created in WFSVM is less than the SVM classifier.

## 1 Introduction

Breast cancer incidence rate is rising in every country of the world especially in developing countries such as India. India is one of the top ten countries in the world having both high incidence and mortality rates according to a report by the International Agency for Research [10] on Cancer. Breast Cancer is rapidly becoming one of the leading Cancers in females with one in 22 women likely to get affected during her lifetime. Breast cancer overall survival rate in India for 5 years is lesser than 60% from reports of Population-Based Cancer Registry indicating a high mortality rate [11]. Though the incidence rate in India is much lesser compared to the western countries, the high mortality rate is due to the lack of instruments and techniques required for the early detection of breast cancer. Early detection refers to tests and exams used to find breast cancer, in people who do not have any symptoms.

One of the best methods used for the detection of breast cancer is mammography. Mammography is the process of using low-dose amplitude X-rays to examine the human breast and is used as a diagnostic and a screening tool. A diagnostic mammogram is used to diagnose breast disease in women who have breast symptoms or an abnormal result on a screening mammogram. Screening mammograms are used to look for breast disease in women who are asymptomatic; that is, those who appear to have no breast problems. But both screening and diagnostic mammograms depend on the radiologist's accuracy in reading the mammograms. On average 21% of breast cancers are missed by radiologists as shown by various reports [13].

In Computer-Aided Diagnosis (CAD), feature extraction transforms the data from the high-dimensional space into lower dimensions and feature selection verifies the importance of selected features through classification techniques. A neural network is a supervised learning method to classify data using hidden layers and constructs weight matrices to represent learning patterns with high computational complexity and slow learning. SVM is also a supervised learning method used for data analysis and pattern recognition which reduces computational complexity and has a faster learning rate. The data analysis performed by SVM can be classification or regression analysis. In traditional SVM the existing kernel functions such as Linear, Polynomial and Radial Basis Function (RBF) are used widely for binary class and multi-class classification.

The recent WFSVM allows us to precompute the kernel matrix and function using the weighted features derived from the actual features of the image or report [5].

In the research of breast cancer diagnosis with classification techniques, considers either the clinical features given by the radiologist report or the specific features extracted from the mammogram images [1], [2]. To analyze this issue, WFSVM classification is implemented with the feature set of statistical, texture and clinical data. From the extracted features the precomputed linear kernel is constructed and trained with weighted features. Then the test samples are validated with the kernel matrix of the test features [4].

This paper elaborates on the following sections. In Section 2, the system design of the proposed methodology is explained. The experimental results and performance of the SVM and WFSVM approach based on the features from image and clinical data are analyzed and tabulated in Section 3. Conclusion and Future work are summarized at the end.

## 2  System Design

In this system, the mammogram images of MIAS dataset are considered for analysis and evaluation in classification. The images are preprocessed using noise removal and background removal techniques and the ROI is identified. From the ROI, the statistical features and texture features are extracted, and clinical features are obtained directly from the dataset. From the feature set of training samples, the weighted features are computed and which is used for the construction of the precomputed kernel in WFSVM. Then the kernel matrix of test features is given as input to the WFSVM classifier for segregating the tumor region as benign or malignant. The system design is shown in Fig. 1

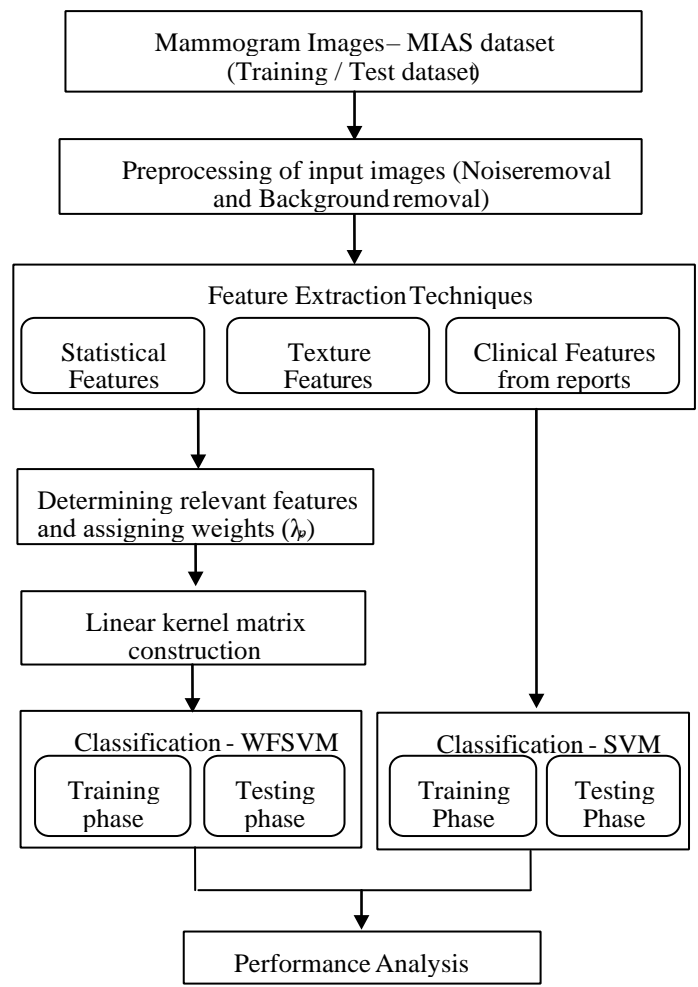

**Fig. 1.** An overall system design

## 2.1 Dataset

The dataset of digital mammogram images is collected from the Mammographic Image Analysis Society (MIAS) database along with the clinical report [9]. It consists of totally 68 benign images and 51 malignant images. For WFSVM and SVM the images are divided equally for both training and testing in each class. In each set of train or test, 58 images are used in which 34 are benign and 24 are malignant.

## 2.2 Preprocessing

Preprocessing techniques such as Noise Removal and Background Removal [15] are applied to mammogram images to extract the required information which leads to efficient classification.

**Noise Removal**: Salt and pepper noise is added to the input image and removed using median filtering. The median filter is a nonlinear digital filtering technique, often used to remove noise. The main idea of the median filter is to run through the signal entry by entry, replacing each entry with the median of neighboring entries. The pattern of neighbors is called "window", which slides, entry by entry, over the entire signal, resulted in a denoised image.

**Background Removal**: The purpose of the cropping step is to focal point the process exclusively on the appropriate breast region, which reduces the possibility for erroneous classification. Hence, the regions of the image which are not part of the breast called artifacts are cropped. On the denoised image, cropping was performed and a cropped image is obtained.

## 2.3 Feature Extraction

The statistical image features such as mean, variance, skewness, uniformity, entropy, kurtosis, contrast, smoothness and the texture features using the average and variance of 6 Gabor filter responses aligned in 30-degree increments are extracted from the ROI of the mammogram images and the clinical features such as the character of background tissue (fatty, fatty glandular, dense glandular) and a class of abnormality present (asymmetry, miscellaneous, architectural distortion, calcification, circumscribed masses, spiculated masses) are obtained directly from the dataset.

**Statistical Features**: The mathematical model to compute the eight statistical features of a mammogram image are listed in Table 1 where $z_i$ is a random variable indicating the intensity, $p(z_i)$ is the histogram of the intensity levels in a region, L is the number of possible intensity levels, σ is the standard deviation [3].

**Table 1.** Statistical Features Formula

| Feature | Formula | Description |
|---|---|---|
| Mean | $\sum_{i=0}^{L-1} z_i * p(z_i)$ | A measure of average intensity |
| Variance | $\sum_{i=0}^{L-1} (z_i - m)^2 * p(z_i)$ | Second moment about the mean. |
| Skewness | $\sum_{i=0}^{L-1} (z_i - m)^3 * p(z_i)$ | Third moment about the mean |
| Uniformity | $\sum_{i=0}^{L-1} p^2(z_i)$ | Measures the uniformity of intensity in the histogram |
| Entropy | $-\sum_{i=0}^{L-1} p(z_i) * \log_2 p(z_i)$ | A measure of randomness of intensity in the histogram |
| Kurtosis | $\sum_{i=0}^{L-1} (z_i - m)^4 * p(z_i)$ | Fourth moment about the mean |
| Contrast | $\sum_{i=0}^{L-1} \sqrt{(z_i - m)^2 * p(z_i)}$ | Standard deviation of pixel intensities |
| Smoothness | $1 - \dfrac{1}{(1+\sigma^2)}$ | Measures the relative intensity variations in a region |

**Texture Features**: Texture features are extracted using the 2-D Gabor filter, as given in Equation (1).

$$h(x,y) = \exp(-\alpha 2\omega(x^2 + y^2)/2) \exp(\omega\pi\alpha\omega(x\cos\theta + y\sin\theta)) \qquad (1)$$

$$\text{where } \alpha = 1/2^{1/2}, \omega = 0,1,2..., \theta = [0, 2\pi]$$

On a trial and error basis, it is found that the filter provides consistent and effective results for the values of ω =2 and θ=5π/3. Different choices of scale and orientation components can be used to construct a set of filters [6]. Here three scales and two orientations are used in the construction of 6 filters. A sample cell array of the Gabor filter bank in three scales and two orientations is 3 x 2 array, computed using the above formula {[1.2682 0.1991] [16.8892 2.6517] [288.7979 45.3932]}. To reduce the computational load, the filter-banks should be made as small as possible in evaluation.

**Convolution**: Once a series of Gabor filters have been chosen, image features at different locations, frequencies, and orientations can be extracted by convolving the image $i(x,y)$ with the filters using the formula in Equation (2)

$$m(x,y) = L_h(i(x,y)) = i(x,y) * h(x,y) \qquad (2)$$

The filter bank is applied to the ROI of an input image and the mean and variance of the filtered image are obtained.

### 2.4 Relevant Features and Assigning Weights

After extracting the low-level features such as statistical, texture, and clinical, the next step is to assign weights to the features that are relevant to classification. Some of the image features are more relevant to the class than the others. It is necessary to identify the relevant features so that the calculation of the kernel function of the support vector machine is not dominated by irrelevant features. The weights of the features are calculated using the technique of principle of maximizing deviations.

**Principle of maximizing deviations:** Consider two classes A and B. The feature vector of a sample that belongs to A and B are given as [7]:
A= ($a_1, a_2, ......a_n$)
B= ($b_1, b_2, ......b_n$)

If the difference between in the $p^{th}$ (p=1,2,….n) feature $a_p$ and $b_p$ of two samples that belong to A and B is more, then that feature plays an important role in classification. Therefore, the feature with greater deviation should be given greater weight than the feature with a smaller deviation. Each feature is a random variable. The deviation of random variables $a_p$ and $b_p$ is given in Equation (3)

$$d(a_p, b_p) = \iint_{-\infty}^{+\infty} |a_p - b_p| f_p(a_p, b_p) da_p db_p \qquad (3)$$

where $f_p(a_p, b_p)$ is the joint probability density function of random variables $a_p$ and $b_p$. The same feature value of different samples are independent, thus we have

$$fp(ap, bp) = fp(ap)\, fp(bp) \qquad (4)$$

Now Equation (3) becomes

$$d(a_p, b_p) = \iint_{-\infty}^{+\infty} |a_p - b_p| f_p(a_p) f_p(b_p) da_p db_p \quad (5)$$

The deviation between categories of samples is given in Equation (6)

$$D(\lambda_p) = \sum_{p=1}^{n} \lambda_p d(a_p, b_p) \quad (6)$$

Structure the model for maximizing the deviation between categories as follows:

$$\max D(\lambda_p) = \sum_{p=1}^{n} \lambda_p d(a_p, b_p) \quad (7)$$

Such that

$$\sum_{p=1}^{n} \lambda_p^2 = 1, \lambda_p \geq 0, p = 1, 2, \ldots n. \quad (8)$$

Using Lagrangian function method, solve the model and the weight of each feature is obtained as given in Equation (9)

$$\lambda_p = \frac{d(a_p, b_p)}{\sum_{p=1}^{n} d(a_p, b_p)} \quad (9)$$

**Precomputed Linear Kernel for WFSVM:** In the precomputed kernel, the kernel values are computed using linear kernel function [14]. The precomputed kernel matrix is used in training and testing files. In that case, the SVM does not need the original training and testing files. Assume there are L training instances $x_1$, $x_2$, …$x_L$. Let K(x,y) be the kernel value of two instances $x$ and $y$. The input formats of training and testing files are:
New training instances for $x_i$: <label> 0:$i$
1:$K(x_i,x_1)$……………L:$K(x_i,x_L)$ New testing instances for any x:
? 0:1 1:$K(x,x_1)$……………L:$K(x\ x_L)$

That is, in the training file the first column must be the class label of $x_i$. In testing, ? can be any value. All kernel values including zeros must be explicitly provided. Any permutation or random subsets of the training/testing files are also valid. The calculation of the precomputed kernel is explained with an example.

***Example:*** Assume the original training data has three instances ($x_1$, $x_2$, and $x_3$) with four features and testing data has one instance.

Training set:
    10    1:1    2:1    3:1    4:1
    40          2:3         4:3
    20               3:1

Test set:
    10    1:1        3:1

The linear kernel, $K(x_i, x_j) = x_i^T \cdot x_j$ is used, to calculate the new training/test sets:
    10    0:1    1:$K(x_1,x_1)$    2:$K(x_1,x_2)$    3:$K(x_1,x_3)$
    40    0:2    1:$K(x_2,x_1)$    2:$K(x_2,x_2)$    3:$K(x_2,x_3)$
    20    0:3    1:$K(x_3,x_1)$    2:$K(x_3,x_2)$    3:$K(x_3,x_3)$

The individual parameters of kernel matrix for the instance $x_1$ are computed as:
    $K(x_1,x_1) = 1*1 + 1*1 + 1*1 + 1*1 = 4$
    $K(x_1,x_2) = 1*0 + 1*3 + 1*0 + 1*3 = 6$
    $K(x_1,x_3) = 1*0 + 1*0 + 1*1 + 1*0 = 1$

Similarly other instances are also calculated. The constructed Kernel matrix using linear kernel for all the three instances of the training set is:
    10    0:1    1:4    2:6    3:1
    40    0:2    1:6    2:18    3:0
    20    0:3    1:1    2:0    3:1

For SVM without weights the kernel matrix constructed above is used in training and testing as it is but for WFSVM the diagonal of the kernel matrix i.e. $K(x_1,x_1)$, $K(x_2,x_2)$ and $K(x_3,x_3)$ are replaced with the weights of the features calculated using Equation (9).

## 2.5 Classification

The classification of breast tissue type is implemented and validated with:
1) Traditional SVM using different kernel types
2) WFSVM using precomputed Linear Kernel with relevant weights in the diagonal of the kernel matrix and without weights substitution.

For both approaches, the difference lies in choosing the kernel function or constructing the kernel function only. But the Binary SVM classification algorithm given below remains the same [8]. The classification of images into their category includes two phases: the training phase and the testing phase.

**Training Phase**: In this phase, from the training images, the low-level features are extracted with these clinical features are added. For WFSVM, the Linear kernel matrix is constructed with weights / without weights and its model file is created. For traditional SVM, the model file is created from the features for specific kernel types (Linear, Polynomial and RBF).

*Algorithm – Binary SVM*
Step 1: Input sample set T = { $(x_i, y_i)$ } $i=1$ to l where $x_i$ is the feature vector and $y_i$ is the classes.
Step 2: Construct the kernel matrix using the features.
Step 3: Select the appropriate penalty parameter and a positive component.
Step 4: Structure the decision function using Equation (10)

$$f(x) = sgn(\sum_{i=1}^{l} y_i \alpha_i * K(x_i, x) + b^*) \qquad (10)$$

where $b^*$ is the positive component.

The algorithm for SVM based on the weighted feature (WFSVM) [5] is the same as traditional SVM but the diagonal of the kernel matrix is replaced with the weights of the features.

**Testing Phase**: The trained WFSVM and SVM are tested with the features of the test set. From the testing set of MIAS database features are extracted, the kernel matrix is constructed without weights in the diagonal and then given as input to the WFSVM for classification. For SVM the test features are given directly to validate the classification with different kernel types.

## 2.6 Performance Evaluation

The performance of the system is measured using the quantitative metrics such as Sensitivity, Specificity and Accuracy as given in Equations (11)-(13).

$$\text{Sensitivity} = TP / (TP + FN) \qquad (11)$$

$$\text{Specificity} = TN / (TN + FP) \qquad (12)$$

$$\text{Accuracy} = (TP + TN) / (TP + TN + FP + FN) \qquad (13)$$

where
TP (True Positive)  - correctly classified positive cases.
TN (True Negative)  - correctly classified negative cases.
FP (False Positive)  - incorrectly classified as negative cases. FN (False Negative)
- incorrectly classified positive cases.

## 3 Results and Performance Evaluation

The result of classification techniques such as SVM and WFSVM are analyzed for the feature set of statistical, texture and clinical features with quantitative metrics. The mammogram images are a kind of X-ray image with a size of 1024*1024 pixels with 256 level grayscale. The sample image is shown in Fig. 2.

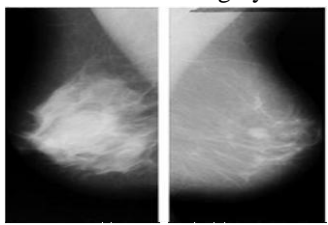

Benign    Malignant

**Fig. 2** Mammogram image classes

Pre-processing techniques such as noise removal and background removal are applied on the mammogram images and the resultant images are shown in Fig 3.

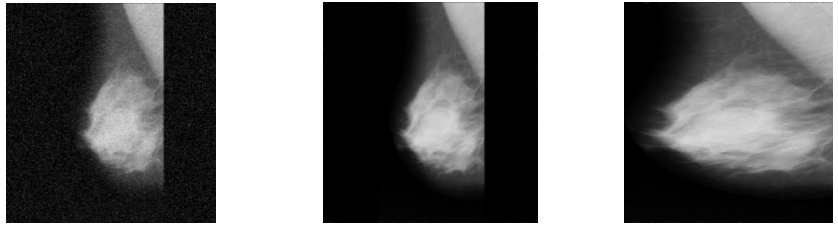

**Fig. 3** Preprocessing – Noise removal and Background removal

After preprocessing, eight statistical features and twelve texture features are extracted from the preprocessed image, and clinical report features are added directly to form a feature set for classification. From this feature space, SVM and WFSVM are performed.

For traditional SVM, the training feature space is used to build a train model file and then the test features are validated. This analysis is carried out for different kernel functions where an equal number of samples are used in training and testing for both the classes. The training is carried out for different feature spaces with various kernel types by changing the *c, g* and *e* values and tested with the test features, resulted in values are given in Table 2. From these results, it has been inferred that texture features are resulting in good accuracy than the other features.

**Table 2.** Performance Analysis of Support Vector Machine with Feature type

| Feature type | Kernel type | Testing(58) misclassified samples | | Accuracy in % | Number of support vectors |
|---|---|---|---|---|---|
| | | B(34) | M(24) | | |
| Statistical features | Lin | 0 | 22 | 60 | 48 |
| | Poly | 0 | 24 | 58 | 53 |
| | RBF | 0 | 24 | 58 | 51 |
| Texture features | Lin | 0 | 1 | 98 | 14 |
| | Poly | 0 | 2 | 96 | 14 |
| | RBF | 0 | 2 | 96 | 10 |
| Statistical, Texture and Clinical features | Lin | 0 | 6 | 90 | 8 |
| | Poly | 0 | 8 | 86 | 8 |
| | RBF | 0 | 11 | 81 | 39 |

Lin– Linear, Poly– Polynomial, RBF– Radial Basis Function B – Benign, M – Malignant.

For WFSVM, the relevant features are identified and maximum weight is assigned using the principle of maximizing deviations between the classes. The precomputed kernel matrix is constructed from the features of the training dataset and the diagonal is replaced with the weighted features for training, from which the train model file is created. Then the kernel matrix is constructed from the features of the test dataset is given as input to the train model file and the classification accuracy is tested. In this approach, to prove the importance of relevant features the constructed kernel matrix is validated without substituting the weights in the diagonal also. In both cases, an equal number of samples is used in training and testing. The detailed result of various analyses is tabulated in Table 3.

**Table 3.** Performance Analysis of Weighted Feature Support Vector Machine with Feature type

| Feature type | Kernel type(4) – Precomputed linear kernel | Testing(58)- misclassified samples | | Accuracy in % | No of support vectors |
|---|---|---|---|---|---|
| | | B(34) | M (24) | | |
| Statistical features | With weights | 0 | 24 | 58 | 44 |
| | Without weights | 1 | 21 | 62 | 51 |
| Texture features | With weights | 0 | 6 | 90 | 2 |
| | Without weights | 0 | 13 | 78 | 12 |
| Statistical, Texture and Clinical features | With weights | 0 | 8 | 86 | 14 |
| | Without weights | 1 | 10 | 81 | 8 |

B – Benign, M – Malignant.

From the results of Table 3, the Precomputed kernel with weights of texture features have resulted in high classification accuracy with less number of support vectors was inferred.

In Table 2 and Table 3, the misclassifications of Benign and Malignant classes are given. i.e. False Negative and False Positive. From these values, True Positive and True Negative can be calculated. The quantitative metrics are analyzed for texture features and shown in Table 4. For analysis, texture features is selected since it has resulted in high accuracy than other combinations. The value of sensitivity indicates that Benign samples are classified correctly but specificity indicates the more misclassifications in Malignant. The reason for this variation in specificity is the number of malignant cases taken for testing is less since large numbers of patient records are not available in the MIAS dataset.

**Table 4.** Performance Metrics Analysis for Texture Features

| Approach used | Accuracy | Sensitivity | Specificity | No of support vectors |
|---|---|---|---|---|
| SVM – Linear | 98 | 100 | 96 | 14 |
| WFSVM- With weights | 90 | 100 | 75 | 2 |
| WFSVM-Without weights | 78 | 100 | 46 | 12 |

**4 Conclusions and Future Work**

In this paper, the efficiency of segregating the mammogram tissue as Benign or Malignant is analyzed using two approaches, such as WFSVM and Traditional SVM with statistical, texture and clinical features. The character of background tissue that dominates the benign and malignant classes are also analyzed and it is inferred that the malignant class falls majorly under the background tissue type dense glandular with tissue shape calcification. In real-time applications, when a single patient record is given as input for testing, then it is classified efficiently using WFSVM since relevant features are assigned with more weights than the other features, whereas in traditional SVM all features are assigned with equal weights. To justify the need for WFSVM, the system was tested without updating the weight values in the diagonal also, which yields less accuracy in classification. Considering the feature set, texture features have resulted in high accuracy in traditional SVM and WFSVM with less number of support vectors. The 2-D Gabor filter of textures features differentiated the tissue type efficiently. In analyzing the statistical features, differentiating the tissue type failed in many cases, since only the grey level values of the pixels are considered in the evaluation for these features.

In future research, the approach used can be tested with DDSM (Digital Database for Screening Mammography) dataset where more patient records are available. The WFSVM can be constructed for other kernel types also. This project can be expanded by incorporating other learning techniques to achieve the task of ensemble learning and understanding in mammogram images